\documentclass[runningheads]{llncs}

\usepackage{xcolor}
\usepackage{graphicx}
\graphicspath{{/}{fig/}}
\usepackage{cite}
\usepackage{tikz}
\usepackage{array}
\usepackage{textcomp}
\usepackage{xcolor}
\usepackage{multirow}
\usepackage{booktabs}
\usepackage{algorithm}
\usepackage{algpseudocode}
\usepackage{subcaption}
\usepackage{svg}
\usepackage{float}

\usepackage{tikz}

\begin{document}
\title{Simulation Analysis of Exploration Strategies and UAV Planning for Search and Rescue}
\author{
    Phuoc Nguyen Thuan\inst{1} \and
    Jorge Peña Queralta\inst{1} \and
    Tomi Westerlund\inst{1}
}%
\authorrunning{P. Nguyen Thuan et al.}
\titlerunning{Simulating UAV Planning Strategies for Search and Rescue}
\institute{
        Turku Intelligent Embedded and Robotic Systems \\
        University of Turku, Finland\\ 
        \email{\{tpnguy, jopequ, tovewe\}@utu.fi} \\
        \url{https://tiers.utu.fi}
    }
\maketitle
%
%
%
\begin{abstract}

Aerial scans with unmanned aerial vehicles (UAVs) are becoming more widely adopted across industries, from smart farming to urban mapping. An application area that can leverage the strength of such systems is search and rescue (SAR) operations. However, with a vast variability in strategies and topology of application scenarios, as well as the difficulties in setting up real-world UAV-aided SAR operations for testing, designing an optimal flight pattern to search for and detect all victims can be a challenging problem. Specifically, the deployed UAV should be able to scan the area in the shortest amount of time while maintaining high victim detection recall rates. Therefore, low probability of false negatives (i.e., high recall) is more important than precision in this case. To address the issues mentioned above, we have developed a simulation environment that emulates different SAR scenarios and allows experimentation with flight missions to provide insight into their efficiency. The solution was developed with the open-source ROS framework and Gazebo simulator, with PX4 as the autopilot system for flight control, and YOLO as the object detector.

\keywords{
    Robotics                            \and 
    Unmanned Aerial Vehicle (UAV)       \and 
    Search and Rescue (SAR)             \and 
    Machine Learning (ML)               \and
    Deep Learning (DL)                  \and
    Active Vision                       \and
    Autonomous Robots
}

\end{abstract}

\section{Introduction}
\label{sec:intro}

The adoption rate of Unmanned Aerial Vehicles (UAVs) has increased in recent years thanks to higher availability and reliability. They have been critical in various civil applications, ranging from surveillance to aiding quick response teams~\cite{shakhatreh2019unmanned}. Many aerial scanning applications aim to build a model of the environment, be it infrastructure~\cite{li2019applications} or mapping a given area with photogrammetry~\cite{krvsak2016use}. However, the purpose of scanning a given area is very often to search for specific objects or people in danger~\cite{queralta2020autosos}. In these cases, the core objective of the flight mission is not necessarily to build an accurate digital representation of the environment but to maximize the probability of finding the people or objects of interest. Particularly, UAVs have shown to be highly efficient in assisting search and rescue (SAR) missions, raising situational awareness among SAR personnel. UAV-based SAR aid systems utilize various sensors, from standard RGB cameras to LIDAR scanners, to provide information about the surroundings and identify and locate missing people or those needing rescuing~\cite{queralta2020sarmrs}.

\begin{figure}[t]
    \centering
    \includegraphics[width=\textwidth]{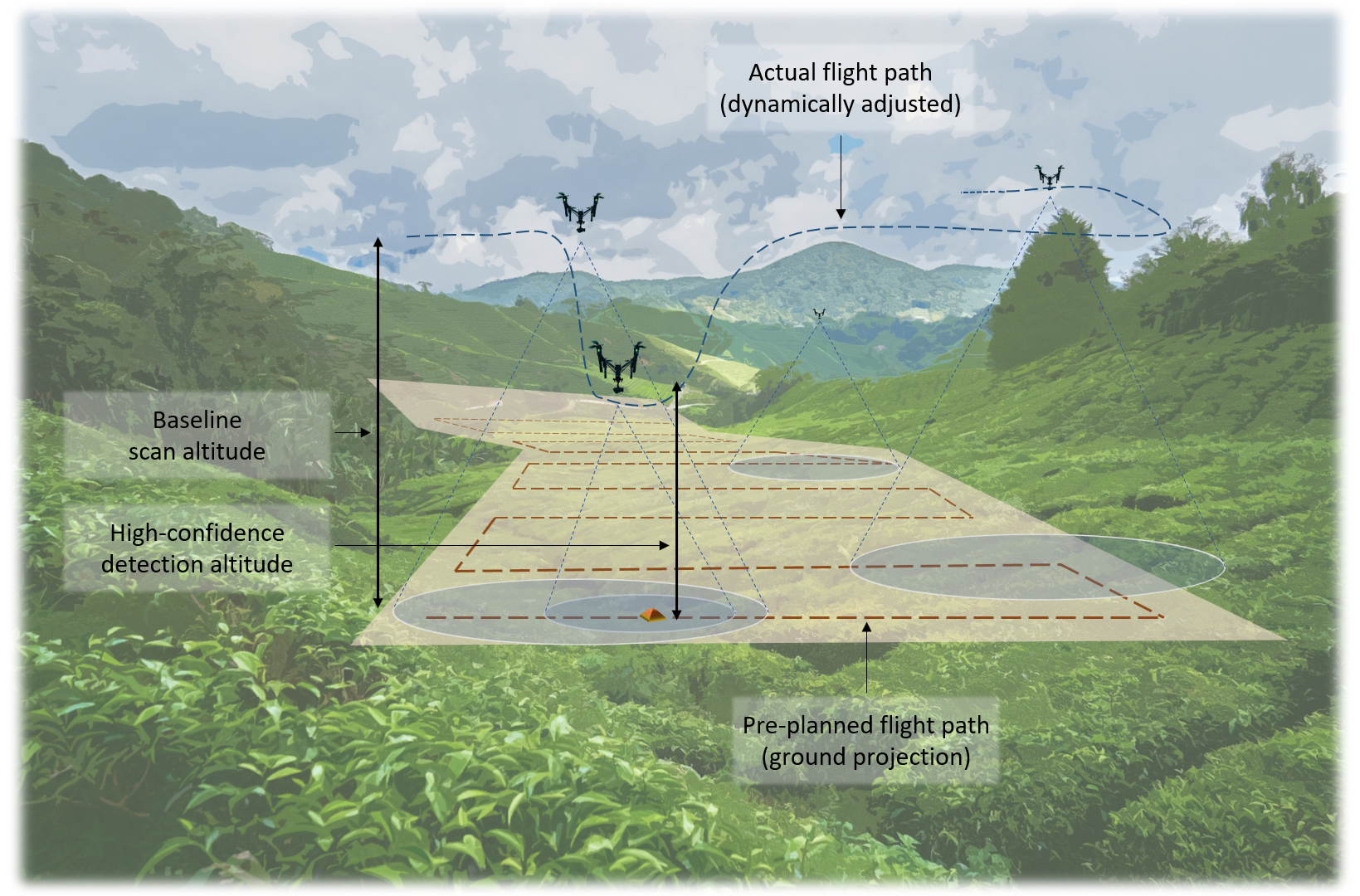}
    \caption{Illustration of a potential flight path followed for searching an object of interest in a given area. The drone altitude varies when areas need to be analyzed in more detail.}
    \label{fig:concept}
    \vspace{22pt}
\end{figure}

This chapter seeks to improve the optimization of aerial search strategy and the respective path planning algorithms for minimizing flight time while ensuring a high recall probability of detecting objects of interest. Automatically-generated path planning algorithms for aerial scans are already standardized nowadays and are usually included in many open-source and commercial software packages~\cite{cabreira2019survey}. These algorithms, however, mainly consider planning paths in two-dimensional space at a certain height. In aerial scans for SAR missions, there is a clear trade-off between the vehicle's altitude and the results' quality. While flying at higher altitudes results in better area coverage and shorter flight time, the results' uncertainties are usually high for most available object detection algorithms and camera hardware. Conversely, detection results for lower altitude flights are often of higher quality. However, flying continuously at such altitudes is time-consuming due to the longer time required to cover broad areas. A hybrid approach combining high altitude for area mapping and scanning and low altitude for verifying the detected object (see Fig.~\ref{fig:concept}) would be a more optimal choice for SAR applications.

A significant challenge with verifying these strategies' efficiency is that experiments with real UAVs in SAR scenarios are challenging to set up due to technical and legislative reasons. Moreover, there are a large number of parameters that can have an impact on the operation, as well as the many different SAR scenarios available. In these situations, simulated experiments are invaluable and can provide insights into the performance of the systems being designed while saving a tremendous amount of time and cost. In this chapter, we describe our approach to designing a customizable simulation environment base on the open-source projects Robot Operating System (ROS)~\cite{quigley2009ros}, Gazebo~\cite{koenig2004gazebo}, and PX4~\cite{meier2015px4}, to emulate different SAR scenarios and experiment with different search strategies. To enable the vehicle to switch flight altitude adaptively, we have integrated an active vision approach into our system~\cite{qingqing2020towards}.

The rest of the chapter is organized as follows. Section \ref{sec:background} overviews the main software foundations that the project leverages. We then describe our early implementation in Section \ref{sec:initim}, with the integration of a deep learning model for object detection and present the autonomous flight algorithm. Section \ref{sec:initres} reports our experimental results, and Section \ref{sec:conc} concludes the work. 
\section{Background}
\label{sec:background}

\subsection{Robot Operating System (ROS)}

ROS is an open-source middleware robotics software platform~\cite{quigley2009ros}. It provides numerous services, ranging from low-level hardware abstraction to message-passing for communication between multiple processes, in the form of software libraries known as packages. For many years, it has facilitated and accelerated the development of robotic software applications in various domains, including industrial robot manipulators, autonomous ground vehicles, and unarmed aerial vehicles. The framework is designed to maximize code reusability for sharing and distributing between different fields of robotic research and development, which has proven to be essential as robot hardware continues to grow in scale and variability.

\subsection{PX4}

PX4 is an open-source autopilot flight stack software for UAVs and other unmanned vehicles~\cite{meier2015px4}. Its many features include full control support for multiple types of aircraft frames, deployment on various flight controller boards, including Linux computers, and multitudes of flexible flight modes for different use cases, including offboard mode specifically for autonomous flights. As a result, it has become the foundation for numerous research and industrious projects in fields including, but not limited to, autonomous flight pathing, aerial tracking, and monitoring. 

\subsection{Robot simulator}

Some popular simulators that support UAVs include X-plane~\cite{xplane}, FlightGear~\cite{perry2004flightgear}, JMavSim~\cite{jmavsim}, AirSim~\cite{shah2018airsim} and Gazebo~\cite{koenig2004gazebo, mccord2019distributed}. Many of these simulators utilize powerful 3D engines to emulate real drones' behaviors. However, for this project, a general robot simulator with support for ROS and different sensors is more desirable than a realistic flight simulator, as it would allow the development of more complex functions and integration with other project components, including the object detector. Therefore, a physics-based simulator is preferable to evaluate flight times and different UAV control and path planning strategies. From the available options, Gazebo is the most straightforward choice with mature ROS and PX4 integration (both software-in-the-loop and hardware-in-the-loop), while it also enables the addition and modifications of robotic sensors~\cite{hentati2018simulation}. As a result, we utilize Gazebo as the main platform to study and experiment with drone flight patterns in this project.
\section{Initial implementation}
\label{sec:initim}

\begin{algorithm}[t]
    \caption{Algorithm to switch between scanning at high altitude and checking at low altitude to collect targets in need of rescuing}
    \label{alg:main}
    \textbf{Input} Predetermined set of waypoints to cover during the flight $F$. Choose flight altitudes $h_S$, $h_C$, the number of targets to scan continuously before switching to checking mode $n_c$, the maximum distance the vehicle can travel from the last target before switching to checking mode $d_{c}$, and the confidence threshold $c_{t}$\\
    \textbf{Output} list of coordinates of targets to rescue $R$
    \begin{algorithmic}
        \State $S \gets []$ \Comment{list of targets discovered while scanning}
        \State $C \gets []$ \Comment{list of targets to check}
        \State $R \gets []$ \Comment{list of confirmed targets in need of rescuing}
        \While{$F$ is not empty}
        \If{$mode$ is SCAN}
            \If{detects $target$}
                \State $handle\_scanned\_target(S, target)$
                \State $last\_target \gets local\_position$
                \State $dist\_traveled \gets 0$
            \Else
                \State $dist\_traveled \gets DIST\left(last\_target, local\_position\right)$
            \EndIf
            \If{$dist\_traveled \geq d_{c}\ OR\ S.size \geq n_c$}
                \State $mode \gets CHECK$
                \State $dist\_traveled \gets 0$
                \State $C \gets S$
            \EndIf
            \If{$local\_position$ == $F[0]$}
                \State $pop(F)$
            \EndIf
            \State $wp \gets \{F[0], h_S\}$
        \ElsIf{$mode$ is CHECK}
            \If{detect $target$}
                \State $handle\_checked\_target(C, target)$
            \EndIf
            \If{$local\_position$ == $C[0]$}
                \State $pop(C)$
            \EndIf
            \If{$C$ is empty}
                \State $mode \gets SCAN$
            \EndIf
            \State $wp \gets \{C[0], h_C\}$
        \EndIf
        \State $move\_to\_waypoint(wp)$
        \EndWhile
        \State \Return $R$
    \end{algorithmic}
\end{algorithm}

\subsection{Object detector}

As our system is modular, many object detection models can be selected as long as the detection results can be formalized into ROS topics. We have performed most of our experiments with YOLOv7~\cite{wang2022yolov7} as the object detector for the system, but integration with other versions of YOLO such as YOLOv4 \cite{bochkovskiy2020yolov4} and YOLOv6 \cite{li2022yolov6} are also possible. The main advantage of the YOLO models over other object detection models is their state-of-the-art performance in real-time situations. In the simulation environment, we used objects present in the COCO dataset~\cite{lin2014microsoft}, which the YOLO models are already pretrained on, as targets. To transition from a simulated environment to real applications, the model can be fine-tuned~\cite{torrey2010transfer} on a different dataset to perform more specialized detection tasks. For instance, in~\cite{qingqing2020towards} we demonstrated how a general pretrained network can be fine-tuned on a more specialized dataset. Specifically, a dataset consisting of aerial images of humans submerged in still water taken at varying altitudes from 20 m to 120 m was collected and used as input to train YOLOv3 for human detection in maritime environments~\cite{redmon2018yolov3}. In general, the algorithms and search strategies that we propose in this chapter are not tailored to any specific object, and therefore we can choose simpler objects for benchmarking the search strategies. Detecting persons from a UAV camera, for example, would require tailoring as most models are trained with frontal images of persons.

\subsection{Flight controller}
\label{subsec:flightcont}

As mentioned in Section~\ref{sec:intro} and~\ref{sec:background}, the main flight controller is developed based on the PX4 framework. The UAV flies in \textit{offboard} mode, which means it operates autonomously without input from any controller. The vehicle's sensor readings and the object detector's outputs are published as ROS topics. The flight path to cover the area of interest is generated using QGroundControl ground station software, which is then used as the main scanning path. During the flight, if a target of interest is spotted, the algorithm will adapt according to some predetermined parameters. These parameters are the number of targets to cluster into one group before checking, denoted by $n_c$, and the travel distance from the previous target before switching to check at low altitude, denoted by $d_{c}$.

Consequently, these two parameters impact the delay between each scanning and checking operation and should be tuned appropriately depending on different scenarios. The control algorithm is formalized as Algorithm~\ref{alg:main}. Other parameters that can impact the performance of the operation are the altitudes when scanning and checking, which are chosen based on the detector's performance at such distance from the targeted object.
\begin{figure}[t]
    \centering
    \includegraphics[width=\textwidth]{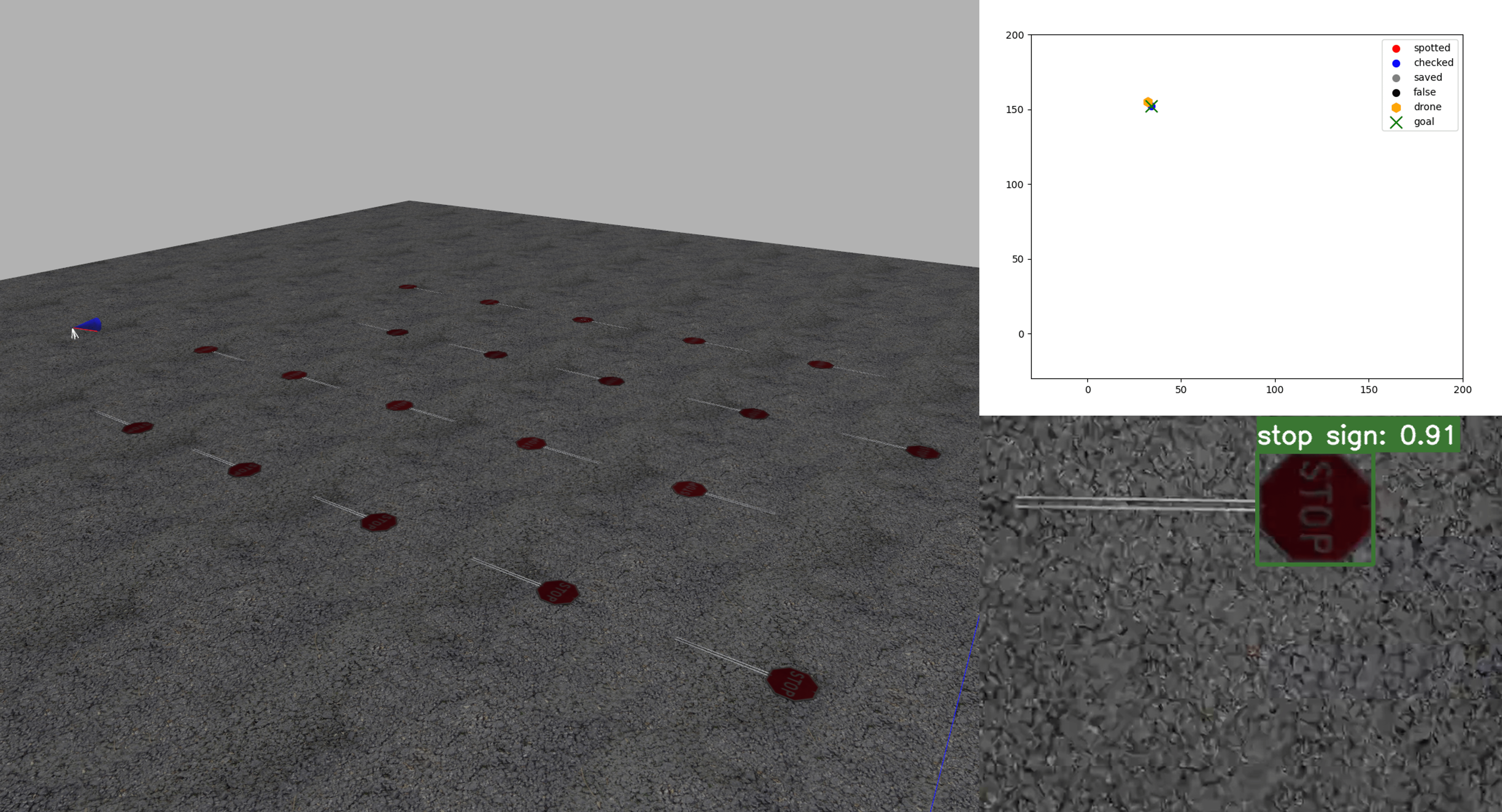}
    \caption{Simulation environment interface. Left: Gazebo simulation environment Top right: debugging map. Bottom right: visualization of object detector output.}
    \label{fig:simenv}
    \vspace{22pt}
\end{figure}

\section{Initial results}
\label{sec:initres}

\begin{table}
    \caption{Comparison between different flight strategies in the environment with 2 clusters of targets}
    \label{table:diffstrat}
    \centering
    \begin{tabular*}{\linewidth}{@{\extracolsep{\fill}} ccc }
        \hline
        Strategy& Mission time (seconds)&False Positives\\
        \hline
        $n_c=1$ & 379.66 & 2\\
        $n_c=5$ & 348.78 & 0 \\
        $n_c=\infty$ & 384.23 & 1\\
        \hline
    \end{tabular*}
\end{table}

\begin{table}
    \caption{Comparison between different flight strategies in the environment with an abundance of targets}
    \label{table:diffstratabun}
    \centering
    \begin{tabular*}{\linewidth}{@{\extracolsep{\fill}} ccc }
        \hline
        Strategy& Mission time (seconds)&False Positives\\
        \hline
        $n_c=1$ & 799.00 & 2\\
        $n_c=5$ & 869.61 & 0 \\
        $n_c=\infty$ & 834.87 & 1\\
        \hline
    \end{tabular*}
\end{table}

\begin{table}
    \caption{Comparison between different flight strategies in the environment with sparsely distributed targets}
    \label{table:diffstratsparse}
    \centering
    \begin{tabular*}{\linewidth}{@{\extracolsep{\fill}} ccc }
        \hline
        Strategy& Mission time (seconds)&False Positives\\
        \hline
        $n_c=1$ & 379.71 & 2\\
        $n_c=\infty$ & 316.61 & 0\\
        \hline
    \end{tabular*}
\end{table}

\begin{figure}
    \centering
    \begin{subfigure}{0.49\textwidth}
      \centering
      \includegraphics[width=\linewidth]{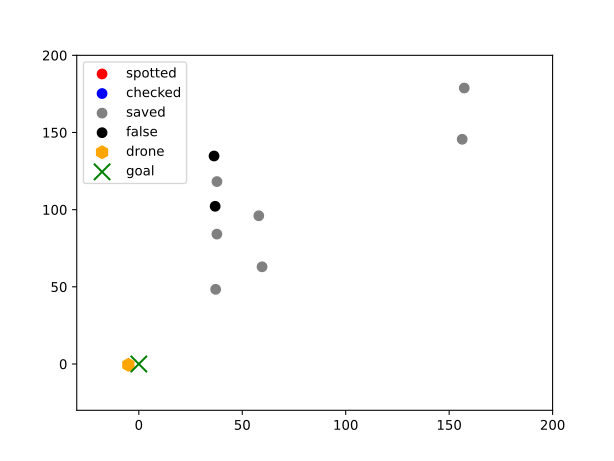}
      \caption{$n_c=1$}
      \label{fig:stratnc1clus}
    \end{subfigure}
    \hfill
    \begin{subfigure}{0.49\textwidth}
      \centering
      \includegraphics[width=\linewidth]{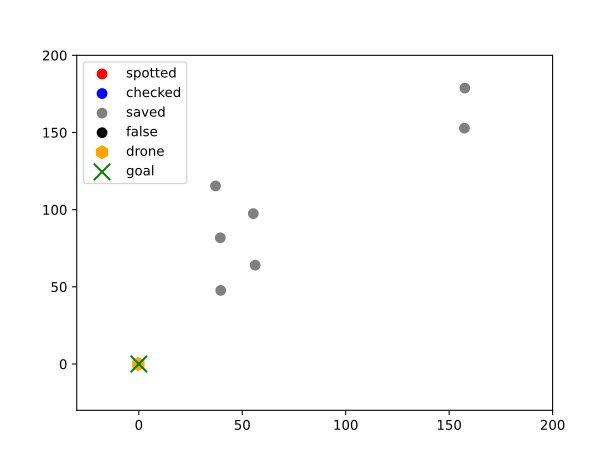}
      \caption{$n_c=5$}
      \label{fig:stratnc6clus}
    \end{subfigure}
    \hfill
    \begin{subfigure}{0.49\textwidth}
      \centering
      \includegraphics[width=\linewidth]{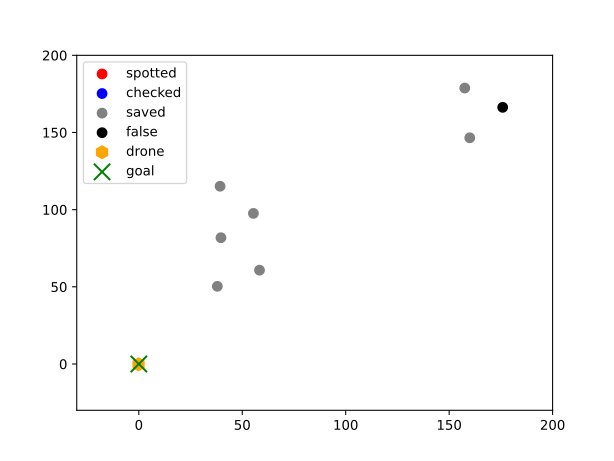}
      \caption{$n_c=\infty$}
      \label{fig:stratncinfclus}
    \end{subfigure}
    \caption{Simulation results from different flight strategies for the environment with 2 clusters of targets.}
    \label{fig:strat_comp_clus}
\end{figure}

\begin{figure}
    \centering
    \begin{subfigure}{0.49\textwidth}
      \centering
      \includegraphics[width=\linewidth]{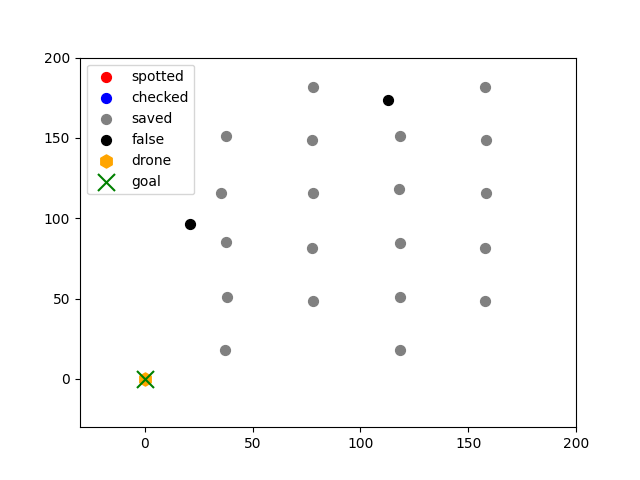}
      \caption{$n_c=1$}
      \label{fig:stratnc1abun}
    \end{subfigure}
    \hfill
    \begin{subfigure}{0.49\textwidth}
      \centering
      \includegraphics[width=\linewidth]{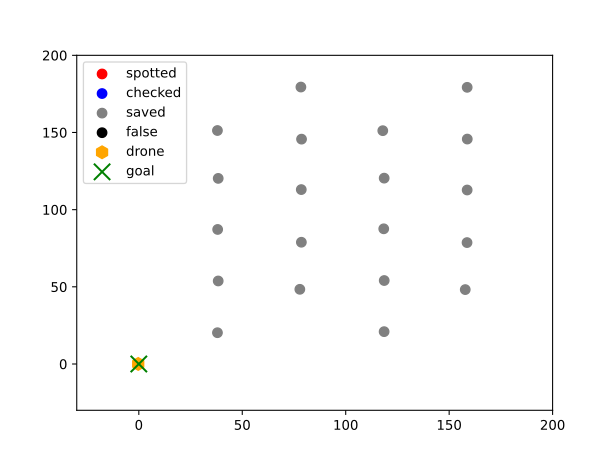}
      \caption{$n_c=5$}
      \label{fig:stratnc6abun}
    \end{subfigure}
    \hfill
    \begin{subfigure}{0.49\textwidth}
      \centering
      \includegraphics[width=\linewidth]{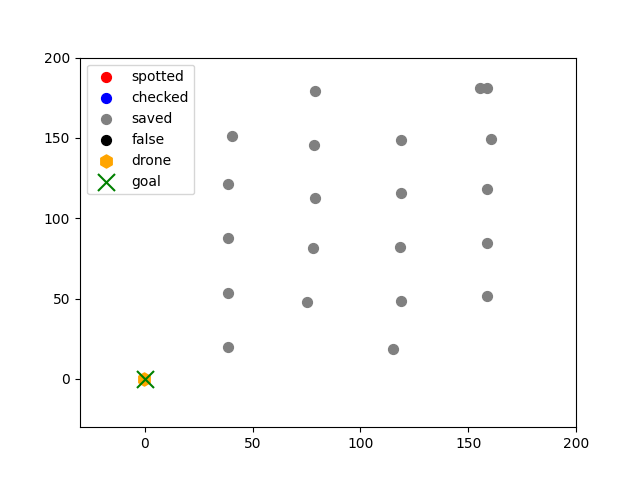}
      \caption{$n_c=\infty$}
      \label{fig:stratncinfabun}
    \end{subfigure}
    \caption{Simulation results from different flight strategies for the environment with an abundance of targets.}
    \label{fig:strat_comp_abun}
\end{figure}

\begin{figure}
    \centering
    \begin{subfigure}{0.49\textwidth}
      \centering
      \includegraphics[width=\linewidth]{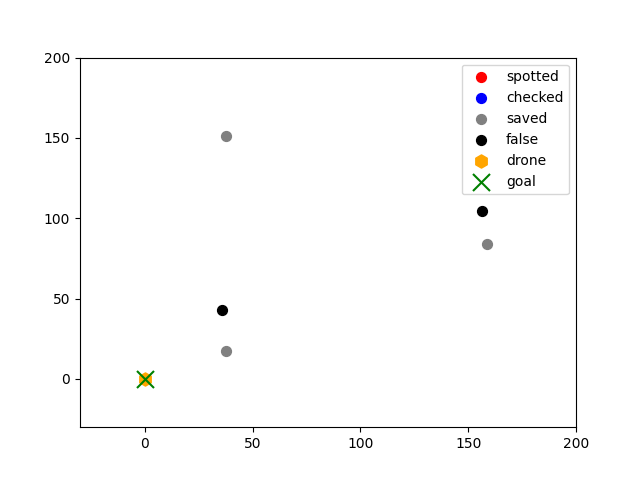}
      \caption{$n_c=1$}
      \label{fig:stratnc1sparse}
    \end{subfigure}
    \hfill
    \begin{subfigure}{0.49\textwidth}
      \centering
      \includegraphics[width=\linewidth]{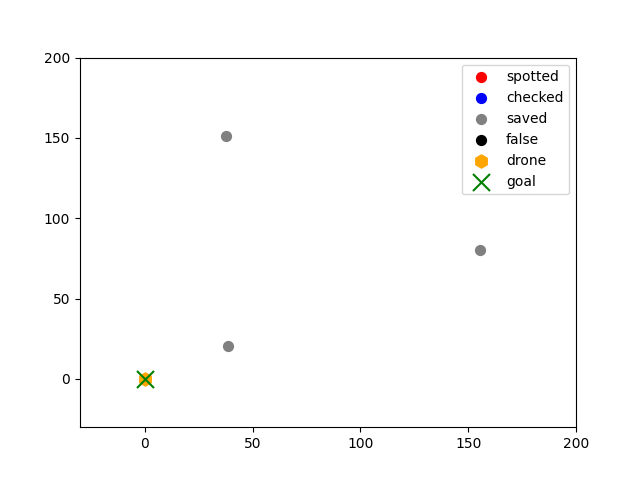}
      \caption{$n_c=\infty$}
      \label{fig:stratncinfsparse}
    \end{subfigure}
    \caption{Simulation results from different flight strategies for the environment with sparsely distributed targets.}
    \label{fig:strat_comp_sparse}
\end{figure}

The Gazebo environment used for the experiments consists of stop signs as targets to search for, and the model being simulated is the Typhoon H480~\cite{gazebovehicles}, with support for video streaming (see Fig.~\ref{fig:simenv}). We performed our experiments in 3 scenarios: one with an abundance of targets, one with 2 clusters of targets, and one with sparsely distributed targets. The parameters determining the delays between finding and inspecting targets, $n_c$ and $d_t$ (see Section \ref{subsec:flightcont}), are varied between 1 to infinity. In the former case, the vehicle will inspect targets as soon as they are discovered, and in the latter case, the vehicle will scan the whole area and return to check on each target individually. The ratio between the speed in scanning and checking flight mode is empirically chosen to be 5 by 1, so that the camera is more stable when validating the detected targets. Tables~\ref{table:diffstrat}, \ref{table:diffstratabun}, \ref{table:diffstratsparse} and Figures \ref{fig:strat_comp_clus}, \ref{fig:strat_comp_abun}, \ref{fig:strat_comp_sparse} provides an in-depth comparison of the drone's performance with different flight strategies in different environments. 

From an early analysis of this data we can conclude that in the cases where the targets are either clustered up or sparse, the strategy that checks every target scanned does not perform as well as the other strategies. From observation, the object detector tends to make false estimations of the targets' locations whenever the vehicle changes flight direction, causing the camera to become unstable. This unwanted behavior is especially costly for this strategy as the drone has to perform multiple redundant vertical movements to check for false targets. However, this performs best when there are many targets to detect because the targets are so close to one another that when the vehicle ascends mid-way, it already detects the next target, thus significantly reducing the number of vertical motions in the mission. In most real situations, this is not the case, as rarely are there cases where multiple targets are lined up in such a way that benefits such strategy. The performance difference between checking targets as clusters and checking every target after one complete scan is insignificant, and both provide their own benefits. Checking targets in clusters reduces the time between the moment a target is scanned and checked, which can especially be beneficial in cases where there are no targets at the end of the mission. However, it is worth emphasizing that the parameters $n_c$ and $d_t$ need to be carefully selected to achieve optimal performance.

\section{Conclusion}
\label{sec:conc}

We have presented and evaluated different search strategies, meaning path planning and active control, for autonomous UAVs. We have done this through the setup of a simulation environment to emulate autonomous UAV flights for search and rescue missions. The environment was constructed based on the foundation of numerous open-source projects, including ROS, PX4, and the Gazebo robot simulator. With the integration of the state-of-the-art YOLOv7 deep neural network for object detection, the simulated vehicle can accurately detect targets in real time and shows potential for sim-to-real transition with transfer learning on more specialized data. Inspired by the work in~\cite{qingqing2020towards} on the effectiveness of object detection models on aerial images at different altitudes, we proposed an autonomous flight algorithm that dynamically switches between high and low altitudes for scanning and checking the targets. Furthermore, through a series of experiments, we have evaluated the effectiveness of different algorithm variations in different scenarios, intending to minimize flight time by reducing unnecessary vertical movements while maximizing the probability of detecting all search targets in the environment.

In future work, we intend to assess the transferability to real-world scenarios as well as introduce new search strategies for multi-UAV systems with active perception approaches.

\section*{Acknowledgment}

This research work is supported by the Academy of Finland's AutoSOS (Grant No. 328755) and AeroPolis (Grant No. 348480) projects.

\bibliographystyle{unsrt}
\bibliography{bibliography}

\begin{thebibliography}{10}

\bibitem{shakhatreh2019unmanned}
Hazim Shakhatreh, Ahmad~H Sawalmeh, Ala Al-Fuqaha, Zuochao Dou, Eyad Almaita,
  Issa Khalil, Noor~Shamsiah Othman, Abdallah Khreishah, and Mohsen Guizani.
\newblock Unmanned aerial vehicles (uavs): A survey on civil applications and
  key research challenges.
\newblock {\em IEEE Access}, 7:48572--48634, 2019.

\bibitem{li2019applications}
Yan Li and Chunlu Liu.
\newblock Applications of multirotor drone technologies in construction
  management.
\newblock {\em International Journal of Construction Management},
  19(5):401--412, 2019.

\bibitem{krvsak2016use}
B~Kr{\v{s}}{\'a}k, P~Bli{\v{s}}t'an, A~Paulikov{\'a},
  P~Pu{\v{s}}k{\'a}rov{\'a}, L'~ml Kovani{\v{c}}, J~Palkov{\'a}, and
  V~Zeliz{\v{n}}akov{\'a}.
\newblock Use of low-cost uav photogrammetry to analyze the accuracy of a
  digital elevation model in a case study.
\newblock {\em Measurement}, 91:276--287, 2016.

\bibitem{queralta2020autosos}
Jorge {Peña Queralta}, Jenni Raitoharju, Tuan~Nguyen Gia, Nikolaos Passalis,
  and Tomi Westerlund.
\newblock Autosos: Towards multi-uav systems supporting maritime search and
  rescue with lightweight ai and edge computing.
\newblock {\em arXiv preprint arXiv:2005.03409}, 2020.

\bibitem{queralta2020sarmrs}
Jorge {Peña Queralta}, Jussi Taipalmaa, Bilge~Can Pullinen, Victor~Kathan
  Sarker, Tuan~Nguyen Gia, Hannu Tenhunen, Moncef Gabbouj, Jenni Raitoharju,
  and Tomi Westerlund.
\newblock Collaborative multi-robot search and rescue: Coordination and
  perception.
\newblock {\em arXiv preprint arXiv:2008.12610 [cs.RO]}, 2020.

\bibitem{cabreira2019survey}
Tau{\~a}~M Cabreira, Lisane~B Brisolara, and Ferreira~Jr Paulo~R.
\newblock Survey on coverage path planning with unmanned aerial vehicles.
\newblock {\em Drones}, 3(1):4, 2019.

\bibitem{quigley2009ros}
Morgan Quigley, Ken Conley, Brian Gerkey, Josh Faust, Tully Foote, Jeremy
  Leibs, Rob Wheeler, Andrew~Y Ng, et~al.
\newblock Ros: an open-source robot operating system.
\newblock In {\em ICRA workshop on open source software}, volume~3, page~5.
  Kobe, Japan, 2009.

\bibitem{koenig2004gazebo}
Nathan Koenig and Andrew Howard.
\newblock Design and use paradigms for gazebo, an open-source multi-robot
  simulator.
\newblock In {\em 2004 IEEE/RSJ International Conference on Intelligent Robots
  and Systems (IROS)(IEEE Cat. No. 04CH37566)}, volume~3, pages 2149--2154.
  IEEE, 2004.

\bibitem{meier2015px4}
Lorenz Meier, Dominik Honegger, and Marc Pollefeys.
\newblock Px4: A node-based multithreaded open source robotics framework for
  deeply embedded platforms.
\newblock In {\em 2015 IEEE international conference on robotics and automation
  (ICRA)}, pages 6235--6240. IEEE, 2015.

\bibitem{qingqing2020towards}
Li~Qingqing, Jussi Taipalmaa, Jorge {Pe\~na Queralta}, Tuan~Nguyen Gia, Moncef
  Gabbouj, Hannu Tenhunen, Jenni Raitoharju, and Tomi Westerlund.
\newblock Towards active vision with {UAV}s in marine search and rescue:
  Analyzing human detection at variable altitudes.
\newblock In {\em IEEE International Symposium on Safety, Security, and Rescue
  Robotics (SSRR)}. IEEE, 2021.

\bibitem{xplane}
X-plane: \url{https://www.x-plane.com}.

\bibitem{perry2004flightgear}
Alexander~R Perry.
\newblock The flightgear flight simulator.
\newblock In {\em Proceedings of the USENIX Annual Technical Conference},
  volume 686, pages 1--12, 2004.

\bibitem{jmavsim}
jmavsim with sitl: \url{https://docs.px4.io/main/en/simulation/jmavsim.html}.

\bibitem{shah2018airsim}
Shital Shah, Debadeepta Dey, Chris Lovett, and Ashish Kapoor.
\newblock Airsim: High-fidelity visual and physical simulation for autonomous
  vehicles.
\newblock In {\em Field and Service Robotics: Results of the 11th International
  Conference}, pages 621--635. Springer, 2018.

\bibitem{mccord2019distributed}
Cassandra McCord, Jorge~Pena Queralta, Tuan~Nguyen Gia, and Tomi Westerlund.
\newblock Distributed progressive formation control for multi-agent systems: 2d
  and 3d deployment of uavs in ros/gazebo with rotors.
\newblock In {\em 2019 European Conference on Mobile Robots (ECMR)}, pages
  1--6. IEEE, 2019.

\bibitem{hentati2018simulation}
Aicha~Idriss Hentati, Lobna Krichen, Mohamed Fourati, and Lamia~Chaari Fourati.
\newblock Simulation tools, environments and frameworks for uav systems
  performance analysis.
\newblock In {\em 2018 14th international wireless communications \& mobile
  computing conference (iwcmc)}, pages 1495--1500. IEEE, 2018.

\bibitem{wang2022yolov7}
Chien-Yao Wang, Alexey Bochkovskiy, and Hong-Yuan~Mark Liao.
\newblock Yolov7: Trainable bag-of-freebies sets new state-of-the-art for
  real-time object detectors.
\newblock {\em arXiv preprint arXiv:2207.02696}, 2022.

\bibitem{bochkovskiy2020yolov4}
Alexey Bochkovskiy, Chien-Yao Wang, and Hong-Yuan~Mark Liao.
\newblock Yolov4: Optimal speed and accuracy of object detection.
\newblock {\em arXiv preprint arXiv:2004.10934}, 2020.

\bibitem{li2022yolov6}
Chuyi Li, Lulu Li, Hongliang Jiang, Kaiheng Weng, Yifei Geng, Liang Li, Zaidan
  Ke, Qingyuan Li, Meng Cheng, Weiqiang Nie, et~al.
\newblock Yolov6: A single-stage object detection framework for industrial
  applications.
\newblock {\em arXiv preprint arXiv:2209.02976}, 2022.

\bibitem{lin2014microsoft}
Tsung-Yi Lin, Michael Maire, Serge Belongie, James Hays, Pietro Perona, Deva
  Ramanan, Piotr Doll{\'a}r, and C~Lawrence Zitnick.
\newblock Microsoft coco: Common objects in context.
\newblock In {\em Computer Vision--ECCV 2014: 13th European Conference, Zurich,
  Switzerland, September 6-12, 2014, Proceedings, Part V 13}, pages 740--755.
  Springer, 2014.

\bibitem{torrey2010transfer}
Lisa Torrey and Jude Shavlik.
\newblock Transfer learning.
\newblock In {\em Handbook of research on machine learning applications and
  trends: algorithms, methods, and techniques}, pages 242--264. IGI global,
  2010.

\bibitem{redmon2018yolov3}
Joseph Redmon and Ali Farhadi.
\newblock Yolov3: An incremental improvement.
\newblock {\em CoRR}, abs/1804.02767, 2018.

\bibitem{gazebovehicles}
Gazebo vehicles:
  \url{https://docs.px4.io/v1.12/en/simulation/gazebo\_vehicles.html}.

\end{thebibliography}

\end{document}